\documentclass[11pt,a4paper]{article}
\usepackage[hyperref]{naaclhlt2019}
\usepackage{times}
\usepackage{latexsym}
\usepackage{graphicx}

\usepackage{url}

\aclfinalcopy % Uncomment this line for the final submission
\title{Visual Understanding and Narration: \\A Deeper Understanding and Explanation of Visual Scenes}

\author{Stephanie M. Lukin\thanks{* Indicates equal contribution}~, Claire Bonial\footnotemark[1]~, and Clare R. Voss \\
  U.S. Army Research Laboratory \\
  Adelphi, MD 20783\\
  {\tt stephanie.m.lukin.civ@mail.mil}}
\date{}

\begin{document}
\maketitle

\section{Introduction}
We describe the task of {\it Visual Understanding and Narration}, in which a robot (or agent) generates text for the images that it collects when navigating its environment, by answering open-ended questions, such as {\it what happens, or might have happened, here?\/}
This task was first explored in \citet{lukin2018pipeline} where humans wrote narratives answering such questions about images taken by a robot (e.g., Fig.~\ref{office}) during our human-robot interaction research \citep{lukin2018scoutbot}. 
The intersection of object identification and text generation has been explored by \citet{das2017visual} and \citet{antol2015vqa}, however these works stop short of inferencing and narration requirements. % of our task. 
\citet{zellers2018recognition} and \citet{goyal2017something} exploit common sense knowledge of stereotypical, human-centric scenarios in individual images and video clips respectively, % These are in comparison to the atypical environments in Fig.~\ref{office},
yet convey their deductions through multiple choice or slot-filling, rather than generating language or narratives.
We briefly survey related current technology and resources, and then sketch our two-pronged approach to bridging the gaps between these fundamental tasks and requirements of the new task.

 %

% 1) single image task, each independent of the others; 
% (so no story / detective work / ideas developed over time; no self-correction)
% 2) no NLG, all multiple choice, first selecting one of four given answers & then selecting one of four given justifications, (they introduce clever Adversarial Matching & also procure nl text via crowdsourcing)
% 3) scene understanding taps into "common sense" knowledge of stereotypical scenarios & of the images shown, they all have people 
% (our images have no people, robot going into possibly anomalous /unusual situations /settings)

%We address two critical NLP and computer vision challenges for enabling a robot to describe its visual surroundings to a remotely located human teammate: 1) what commonsense information is required to reason about objects and how do we provide that information? and 2) how do we select what to talk about and weave together a cohesive narrative?  
%We survey existing resources that address some of these needs and introduce our own approaches: a bottom-up strategy leveraging an ontology of ``qualia'' associated with particular objects, and a top-down strategy leveraging crowd-sourced narratives of images taken onboard a robot (Fig.~\ref{office}) in the course of our human-robot interaction research \citep{lukin2018scoutbot}.

\section{Visual Understanding}

Addressing the gap between recognizing particular objects in images and reasoning about why they may be present in a given physical environment requires commonsense knowledge.
%and how this can best be organized.  
%Related work and existing resources are surveyed briefly below, followed by an introduction to a new resource developed for use in our own scene understanding pipeline.  

\subsection{Commonsense Gaps}
%There is increasing interest in 
Commonsense knowledge about objects for computer vision has
%recently 
shown both to improve object and activity recognition and
to provide additional information necessary for deeper reasoning  \cite{gupta2015visual,yatskar2016situation,ronchi2015describing}.  
This type of object knowledge %prioritized in this research 
is primarily visual, supporting tasks such as object and activity recognition, as well as transfer learning to visually similar objects and scenes.  
Such knowledge has included spatial relations \cite{yatskar2016stating}, shape similarity to other objects, and visual attributes such as color \cite{kumar2018dock}. 
%Yatskar, Ordonez \& Farhadi \shortcite{yatskar2016stating} extract three types of knowledge for objects in the MS-COCO dataset: spatial relations (e.g., (holds(bed, dog))), entailment rules (e.g., (holds(bed, dog) \textit{entails} laying-on(dog, bed)), and generalization to unseen objects using WordNet \cite{fellbaum1998semantic}.  Singh et al. \shortcite{kumar2018dock} also use the MS-COCO dataset and extract ``commonsense knowledge" based on an object’s visual similarity to other objects, similarity of attributes such as color, and spatial relations in order to aid transfer learning and improve object recognition. 
%Such knowledge has been collected via automatic extraction from existing external knowledge sources including ontologies (e.g., Yatskar, Ordonez \& Farhadi \shortcite{yatskar2016stating}), as well as crowdsourced annotations of visual data (e.g., Gupta \& Malik \shortcite{gupta2015visual}; Yatskar, Zettlemoyer \& Farhadi \shortcite{yatskar2016situation}). 
%

\begin{figure}
    \centering
        \includegraphics[width=0.45\textwidth]{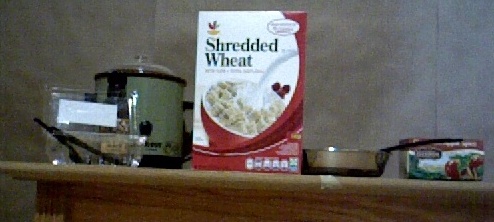}
    \caption{Image taken onboard robot}\label{office}
    \vspace{-0.2in}
\end{figure}

However, knowledge humans exploit when analyzing an environment goes beyond visual clues. 
To interpret Fig.~\ref{office} possibly as a kitchen, a system needs not only to recognize the objects, but to know which actions are commonly performed with these objects, and then to infer where such actions may occur. %Zhu, Fathi \& Fei-Fei \shortcite{zhu2014reasoning} focus specifically on 
	The actions, termed `object affordances' %: the properties of an object that determine the actions a human can perform on them 
	\cite{gibson1979}, have been defined in computer vision studies as the combination of: an affordance label, a human pose representation of the action, and a relative position of the object with respect to the human \cite{grabner2011makes,kjellstrom2011visual,yao2018temporal,zhu2014reasoning}.  Though the latter two can be extracted from visual data, a challenge is how to systematically collect appropriate affordance labels for shared re-use by vision and language researchers, reducing the redundant labor of independent, manual assignments of verbs as labels for small, fixed sets of objects (e.g., \textit{sit-on -- chair}).

\subsection{Approach to Bridging Gaps}
%This fails to take advantage of the linguistic resources from which we can extract a large number of general object affordances and other types of commonsense information. 
`Qualia' are relations associated with a particular object \cite{pustejovsky1991generative}, including Agentive ({\sc created\_by}), Telic ({\sc functions\_as, used\_for}), Constitutive ({\sc part\_of, made\_of}), and Formal ({\sc is\_a}), providing a rich source of commonsense and affordance information, and a
%Within NLP and linguistics research, 
%They provide a 
framework for disambiguating senses of a word (e.g., %She threw the book at him – referring to the 
{\it book:\/} physical item vs. %She wrote the book – referring to 
content). They have been demonstrated as useful knowledge representations for intelligent agents \cite{mcdonald2013representation,pustejovsky2017object,narayana2018cooperating}.
	
A comprehensive set of qualia relations have yet to be defined and organized. We are tackling this challenge and aim to make the qualia usable for visual understanding tasks: qualia have been automatically extracted and evaluated for quality via crowdsourcing \cite{kazeminejad2018automatically}, then encoded as relations between entities and events in the Rich Event Ontology (REO)
% an ontological backbone of events unifying existing textual semantic role labeling resources
 \cite{bonial2016multimodal}.   
%These SRL resources provide the largest datasets of text marked up with semantic roles, or ``who is doing what to whom."  Each resource, however, marks up both event types and semantic roles of varying definitions and specificity; REO is the first and only effort to unify these previously disparate schemas.  While this allows for the combination of these resources into a larger and more diverse set of training data supporting automatic SRL in text, there is also a valuable opportunity to leverage text data for interpreting ``what’s happening" in visual and/or multimodal data. Thus, t
%The qualia, as encoded in REO, represent one-of-a-kind, large-scale, computer-readable knowledge representations potentially allowing an intelligent agent to reason about its visual environment once objects are recognized.
Assuming, for example, that the objects in Fig.~\ref{office} can be recognized accurately, the resulting list of objects (e.g., \textit{pot, cereal}) can first be queried for their qualia in REO, to discover: \textit{pot} is {\sc used\_for} \textit{cooking},  and \textit{cereal} {\sc functions\_as} \textit{nourishing} and {\sc is\_a} \textit{Prepared\_Food}.  These activities with object classes can next be queried for their common locations in REO via their semantic roles, to discover: \textit{cooking} {\it Prepared\_Food} returns \textit{kitchen}. In this approach, the objects, their affordances, and REO roles, would support the inference that this space {\sc functions\_as} as a kitchen. 
%What is pictured in
% Fig.~\ref{office} provides a robot-level view, but a human could guess the space {\sc functions\_as} as a kitchen.
%, and it is valuable that our robot teammates could also. 
%Thus, the qualia encoded in REO provide bottom-up, commonsense object information that can be used for reasoning. % about the environment. % While REO captures typical information about objects that many people assume and may not think to describe (i.e. ``food is for eating"), in the next section, we describe efforts to recognize and provide a narrative about what might be more notable/anomalous to a human observing this environment. 

\section{Narrative Building}

%a.  What do we talk about? (Framing)
%b. How do we talk about it? (Construction)
%c. existing resources pros and gaps, Stephanie's approach

Once the visual scene is interpreted, we determine what is needed to answer the task question via content selection and narrative generation. 

\subsection{Generation Gaps}
Content selection, or framing, is the relationship between the narrating agent and what they know and choose to talk about 
%``accessibility of knowledge needed to select ... events for presentation in discourse'' 
\citep{lonneker2005narratological}.
%This selection of content, also known as aspect, has been studied from a grammatical perspective with respect to recalling events, exploring the differences between asking a question in the imperfect tense (``what was happening'') or in the perfect tense (``what happened''), yielding different selected content from the same visual scene \citep{matlock2012smashing}. 
%both of which has shown that when humans think about actions, they mentally imagine themselves doing so, which is seen in the brain \citep{rizzolatti2008mirrors}, and a similar result is found when people read stories \citep{speer2009reading}. 
%The appropriate framing device can be utilized depending on the intended audience of the final narrative.
The choice of appropriate framing device depends on the intended audience of the final narrative.
Many recent works in vision % frame this 
treat framing as an observational task, describing the image in a single sentence 
\citep{rashtchian2010collecting, hodosh2013framing,lin2014microsoft,chen2015microsoft,krishnavisualgenome}.\footnote{\citet{ferraro2015survey} %conduct a %comprehensive 
survey of %available 
vision and language resources; framing prompts are similar to those listed here.} 
This limits the scope to the visually observable and restricts what can be learned by extrapolation from the past or to future.
% and the exploration of agent's inner states. 
With just a handful of open-ended prompts, e.g., {\it what happened,\/} creative scene interpretations can be elicited that go beyond single sentences  \citep{gordon2014authoring,huang2016visual,vaidyanathan2018snag}.
%Other visual prompts, however, do allow for these kinds of interpretations to form \citep{gordon2014authoring} ask subjects to write ``what happened'' for a 90 second animated film with 2-dimension shapes, and observed that subjects often attributed emotions and backstories to the shapes. \citet{huang2016visual} ask subjects to ``write a sentence or a phrase on each [image] to form a story'' which captures the thread of narrative across multiple images, but is still restrictive to a single sentence. \citet{vaidyanathan2018snag} record subject gaze in real-time as they ``describe the action in the images and tell the experimenter what is happening.''

% b) 
After assessing {\it what} to talk about, the narrating agent must establish {\it how} to talk about it. Recent neural vision and text models rely solely on crowd-sourced data for guidance in this phase of narrative crafting \citep{park2015expressing,yu2017hierarchically,huang2016visual,fan2018hierarchical,wang2018no}. Much can be learned from narratological studies, such as the categorization, combination, and presentation of narrative elements 
%into evaluation, orientation, and action 
\citep{labov1997narrative,rahimtoroghi2013evaluation,niehaus2009computational,lehnert1981plot,elson2012dramabank}.
%or incorporate, inference, understand \citep{niehaus2009computational}.
% Language is generated based on the narrative shape and content. 
However, the template-based approaches \citep{montfort2007generating,callaway2002narrative} and statistical models \citep{li2015learning} that have successfully leveraged these elements for content selection and narrative shaping in text-based story generation, have not yet been applied to visual narration.

\subsection{Approach to Bridging Gaps}
\citet{lukin2018pipeline} performed a pilot data collection with framing %that encouraged consideration of the 
to elicit a narrative connecting a sequence of images. In our ongoing work, preliminary analysis of human authored narratives about Fig.~\ref{office} have found both extrapolation beyond the observable in the image (``[someone intends] to live here at least until they finish the project that they are working on'') and creative causal reasoning for what is not visually depicted in the image (``[someone] is pulling an all-nighter and brought breakfast for the next morning''). 
%Both narratives would be important for a robot to generate as it explores an unknown environment.

\section{Next Steps}
The two prongs of our approach provide complementary information for identifying and reasoning about a visual scene, from which succinct and targeted text can be generated in support of human-robot interactions to talk about what happens in the robot's environment. Qualia encoded in REO provide bottom-up, commonsense knowledge for reasoning, and existing narrative schema can be applied in a top-down manner to formulate narratives, leveraging content from crowd-sourced narrative elements.
% compared to only what is visually observable, and templatic, statistical, and neural language models explored.
%
%REO captures semantic typical information about objects used for reasoning %that people assume 
%that people may not think to describe (i.e. ``food is for eating'') which is used to inform the narrative construction detailing the implications of these observations in a particular environment.
%TBD remove redundancy: We plan to integrate both research trajectories into a scene summarization pipeline to be used onboard a robot in which the ontological component supports some reasoning over objects detected, while the narrative planning component supports weaving together this object information into a cohesive description. 
Our ontology and crowdsourced annotations will be made available to the community, supplementing existing resources.

\bibliography{sivl2019}
\bibliographystyle{acl_natbib}

\end{document}